\documentclass{vgtc}                          

\ifpdf
  \pdfoutput=1\relax                   
  \usepackage{graphicx}                
  \DeclareGraphicsExtensions{.pdf,.png,.jpg,.jpeg} 
\else
  \usepackage{graphicx}                
  \DeclareGraphicsExtensions{.eps}     
\fi%

\graphicspath{{figures/}{pictures/}{images/}{./}} 

\usepackage{microtype}                 
\PassOptionsToPackage{warn}{textcomp}  
\usepackage{textcomp}                  
\usepackage{mathptmx}                  
\usepackage{times}                     
\usepackage{cite}                      
\usepackage{tabu}                      
\usepackage{booktabs}                  
\usepackage{graphicx}
\usepackage[utf8]{inputenc}
\usepackage{algorithm}
\usepackage{algpseudocode}
\usepackage{amsmath}
\usepackage{slashbox}
\usepackage{bm}
\usepackage{amsmath,amssymb,amsfonts}
\usepackage{url}
\usepackage{hyperref}
\usepackage[utf8]{inputenc}
\usepackage{amsthm}
\usepackage{tikz}
\usetikzlibrary{shapes.geometric}
\theoremstyle{remark}

\onlineid{0}

\vgtccategory{Research}

\newcommand{\R}{{\mathbb R}}

\title{
    Kernel t-distributed stochastic neighbor embedding
    \thanks{
        This work was supported by a grant of the Ministry of Research,
        Innovation and Digitization, CNCS - UEFISCDI,
        project number PN-III-P4-PCE-2021-0154, within PNCDI III.
    }
}

\author{Denis C. Ilie-Ablachim\thanks{e-mail: denis.ilie\_ablachim@upb.ro}\\
       \scriptsize University Politehnica of Bucharest
\and Bogdan Dumitrescu\thanks{e-mail: bogdan.dumitrescu@upb.ro}\\
     \scriptsize University Politehnica of Bucharest
\and Cristian Rusu\thanks{e-mail: cristian.rusu@unibuc.ro}\\
     \scriptsize University of Bucharest
}

\abstract{This paper presents a kernelized version of the t-SNE algorithm, capable of mapping high-dimensional data to a low-dimensional space while preserving the pairwise distances between the data points in a non-Euclidean metric.
This can be achieved using a kernel trick only in the high dimensional space or in both spaces, leading to an end-to-end kernelized version.
The proposed kernelized version of the t-SNE algorithm can offer new views on the relationships between data points, which can improve performance and accuracy in particular applications, such as classification problems involving kernel methods.
The differences between t-SNE and its kernelized version are illustrated for several datasets, showing a neater clustering of points belonging to different classes.
}

\CCScatlist{
  \CCScatTwelve{nonlinear dimensional reduction}{kernel methods}{nonlinear mapping}{t-SNE};
}

\begin{document}


\firstsection{Introduction}

\maketitle

Dimensional reduction is the technique of reducing the size of data situated in a high-dimensional space into a low-dimensional one which is more favorable for analysis and visualization. The most important characteristic of the resizing procedure is retaining relevant information about data. The applications of dimensional reduction algorithms can be extended to other problems, such as feature selection, data compression, and computation optimization. 

There are several well-known dimensionality reduction algorithms used in machine learning and data analysis, such as Sammon mapping \cite{sammon}, Isomap \cite{isomap}, Locally Linear Embedding (LLE) \cite{lle}, Spectral Embedding \cite{spectral_embedding}, Principal Component Analysis (PCA) \cite{pca} and its kernelized version Kernel PCA \cite{kpca}. Each method uses pairwise distances between the data points in the high-dimensional space and tries to conserve the pairwise distances in the low-dimensional space while preserving the structure of the data. This is obtained by minimizing different cost functions (discrepancy, divergence) or by following the relationships of data based on a weighted graph.

The most widely used method for the visualization of high-dimensional data is t-distributed Stochastic Neighbor Embedding (t-SNE) \cite{van2008visualizing}. This method presents many advantages, but sometimes it might encounter some limitations regarding the distortion of distances or lack of interpretability. For datasets with nonlinear representation properties, a Euclidean metric might not be effective. 

Our contribution is to show that nonlinear metrics may be better suited for certain dimensional reduction tasks. 
We propose two approaches.
In the first, the kernel metric is used only in the high-dimensional space,
while the Euclidean metric is used in the low-dimensional space;
the method is called Kernel t-SNE.
In the second, both high- and low-dimensional spaces are endowed with
the kernel metric; the method is called End-to-End Kernel t-SNE.

The paper is organized as follows.
In section \ref{sec:t-sne} we recall the essentials of the t-SNE algorithm.
Section \ref{sec:kt-sne} is dedicated to the presentation of the two
versions of our method.
Section \ref{sec:res} presents implementation details and the results
obtained on several datasets, together with some examples.
The last section contains the conclusions and some directions of future work.

\section{t-distributed Stochastic Neighbor Embedding}
\label{sec:t-sne}

t-distributed Stochastic Neighbor Embedding (t-SNE) is a machine learning algorithm used for dimensional reduction and visualization of high-dimensional data. Its main objective is the mapping of a high-dimensional space to a low-dimensional space (usually 2D or 3D) while preserving the dataset properties. The algorithm minimizes the Kullback-Leibler divergence between two probability distributions defined in the two spaces, in which the pairwise similarities are respected. Considering points $x_{i} \in \R^{n}$ in the data space and $y_{i} \in \R^{m}$ (with $m < n$) in the corresponding reduced space, the pairwise similarities are defined by the weighted sum between the two conditional probabilities
\begin{equation}
p_{i j}=\frac{p_{j \mid i}+p_{i \mid j}}{2 n}
\end{equation}
for the high-dimensional space, where
\begin{equation}
p_{j \mid i}=\frac{\exp \left(-\left\|x_i-x_j\right\|^2 / 2 \sigma^2\right)}{\sum_{k \neq i} \exp \left(-\left\|x_i-x_k\right\|^2 / 2 \sigma^2\right)},
\label{eq:pji}
\end{equation}
and the low-dimensional map is defined as
\begin{equation}
q_{i j}=\frac{\left(1+\left\|y_i-y_j\right\|^2\right)^{-1}}{\sum_{k \neq l}\left(1+\left\|y_k-y_l\right\|^2\right)^{-1}}.
\label{eq:qij}
\end{equation}
The dimensional reduction strategy consists in minimizing the sum of the Kullback-Leibler divergences between the conditional probabilities $p_{i j}$ and $q_{i j}$
\begin{equation}
C=K L(P \| Q)=\sum_i \sum_j p_{i j} \log \frac{p_{i j}}{q_{i j}}.
\label{eq:C}
\end{equation}
The problem is solved by computing the partial derivative of the objective function with respect to the low-dimensional vector
\begin{equation}
\frac{\delta C}{\delta y_i}=4 \sum_j\left(p_{i j}-q_{i j}\right)\left(1+\left\|y_i-y_j\right\|^2\right)^{-1}\left(y_i-y_j\right)
\end{equation}
and updating the low-dimensional vectors by following a gradient descent procedure with a momentum term.

\section{Kernel t-SNE}
\label{sec:kt-sne}

\begin{figure}[tbp]
  \centering
    \begin{tikzpicture}[scale=2.5]
      \node[draw, ellipse, fill=cyan, minimum width=2.5cm, minimum height=1cm] (A) at (0,0) {$\mathbb R^{n'}$};
      \node[draw, ellipse, fill=cyan!20!white, minimum width=2cm, minimum height=1cm] (B) at (2,0) {$\mathbb R^{m'}$};
      \node[draw, ellipse, fill=green, minimum width=1.5cm, minimum height=1cm] (C) at (0,-1) {$\mathbb R^{n}$};
      \node[draw, ellipse, fill=green!10!white, minimum width=1cm, minimum height=1cm] (D) at (2,-1) {$\mathbb R^{m}$};
      
      \draw[->] (A) -- (B) node[midway, above, sloped] {E2E Kernel t-SNE};
      \draw[->] (C) -- (A) node[midway, left] {$\varphi$};
      \draw[->] (D) -- (B) node[midway, right] {$\varphi$};
      \draw[->] (C) -- (D) node[midway, above, sloped] {t-SNE};
      \draw[->] (A) -- (D) node[midway, above, sloped] {Kernel t-SNE};
      \node[left] at (-0.1,-0.25) {$n' \gg n$};
      \node[right] at (2.1,-0.25) {$m' \gg m$};
    \end{tikzpicture}
\caption{Diagram of dimensional reduction transformations.}
\label{fig:diagram}
\end{figure}
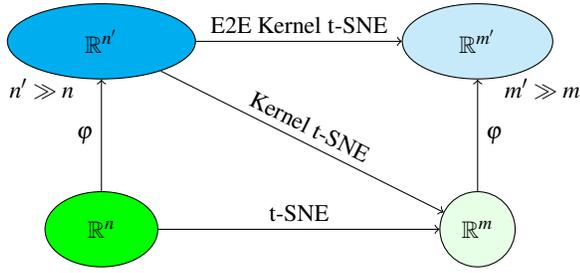

In this section, we present the main contribution of our work. We introduce two dimensional reduction methods, using transformations whose diagram is shown in Figure \ref{fig:diagram}.
Both methods rely on mapping the points $x_{i} \in \R^{n}$ from the data space into the lifted space $\R^{n'}$, with $n' \gg n$, using a nonlinear kernel function $\varphi(\cdot)$.
From the lifted space we can go directly to the reduced space $\R^{m}$ or indirectly, via a lifted reduced space $\R^{m'}$, with $m' \gg m$, related to the reduced space through the same kernel function $\varphi(\cdot)$.
We call (standard) kernel t-SNE the method to produce the former mapping 
and end-to-end kernel t-SNE the method to produce the latter.

\subsection{Standard Kernel t-SNE}

The nonlinear mapping from the data space to the lifted space changes the metric
and thus allows the possibility to capture more complex patterns and to give different interpretations to the decisions boundaries.
Immersing the data points into a higher dimensional space can lead to more significant results.

Accordingly, we redefine the conditional probability \eqref{eq:pji} of the data space by mapping the data points to the lifted space with a function $\varphi(\cdot)$:
\begin{equation}
p_{j \mid i}=\frac{\exp \left(-\left\|\varphi(x_i)-\varphi(x_j)\right\|^2 / 2 \sigma^2\right)}{\sum_{k \neq i} \exp \left(-\left\|\varphi(x_i)-\varphi(x_k)\right\|^2 / 2 \sigma^2\right)}.
\label{eq:pji_phi}
\end{equation}
Since the lifted space is mapped directly to the reduced space, relation \eqref{eq:qij} remains valid.
Relation \eqref{eq:C} between the two conditional probabilities also stays unchanged.
As we will see in the experiments section, the computations have reasonable complexity compared to those of t-SNE.

The original t-SNE algorithm is adapted by only computing in \eqref{eq:pji_phi} the norm $\left\|\varphi(x_i)-\varphi(x_j)\right\|^2$ according to the kernel trick:
\begin{equation}
\left\|\varphi(x_i)-\varphi(x_j)\right\|^2 = \varphi(x_i)^{\top}\varphi(x_i) - 2 \varphi(x_i)^{\top}\varphi(x_j) + \varphi(x_j)^{\top}\varphi(x_j),
\end{equation}
which leads to
\begin{equation}
\left\|\varphi(x_i)-\varphi(x_j)\right\|^2 =k(x_i, x_i) - 2 k(x_i,x_j) + k(x_j, x_j),
\end{equation}
where $k(u,v) = \varphi(u)^{\top}\varphi(v)$ is the proposed kernel function for the nonlinear mapping.

\subsection{End-to-End Kernel t-SNE}

This subsection presents the end-to-end kernel t-SNE, in which the dimensional reduction is achieved by minimizing the Kullback-Leibler divergences between the mappings in the lifted spaces of both data and reduced spaces.

In this case, the lifted version of the low-dimensional conditional probability \eqref{eq:qij} is
\begin{equation}
q_{i j}=\frac{\left(1+\left\|\varphi(y_i)-\varphi(y_j)\right\|^2\right)^{-1}}{\sum_{k \neq l}\left(1+\left\|\varphi(y_k) - \varphi(y_l)\right\|^2\right)^{-1}}.
\label{eq:q_kernel}
\end{equation}
This time we recompute the partial derivative of $C$ with respect to the current reduced sample $y_i$ by taking into account its nonlinear mapping to the kernel space
\begin{equation}
\begin{aligned}
\frac{\partial C}{\partial y_i} &=\sum_j\left(\frac{\partial C}{\partial d_{i j}}+\frac{\partial C}{\partial d_{j i}}\right) \cdot \frac{\partial d_{i j}}{\partial y_i}=\\
&=2 \sum_j \frac{\partial C}{\partial d_{i j}} \cdot \frac{\partial d_{i j}}{\partial y_i}
\end{aligned}
\label{eq:dCdyi}
\end{equation}
where we denoted
\begin{equation}
d_{i j}=\left\|\varphi\left(y_i\right)-\varphi\left(y_j\right)\right\|^2.
\end{equation}
Using the notation
\begin{equation}
Z=\sum_{k \neq l}\left(1+d_{k l}\right)^{-1},
\end{equation}
direct calculus leads to
\begin{equation}
\begin{aligned}
\frac{\partial C}{\partial d_{i j}} &=-\sum_{k \neq l} p_{k l} \frac{\partial \log q_{k l}}{\partial d_{i j}}=-\sum_{k \neq l} p_{k l} \frac{\partial(\log q_{k l} Z - \log Z)}{\partial d_{i j}}=\\
&=-\sum_{k \neq l} p_{k l}\left(\frac{1}{q_{k l} Z} \cdot \frac{\partial\left(1+d_{k l}\right)^{-1}}{\partial d_{i j}} - \frac{1}{Z} \cdot \frac{\partial Z}{\partial d_{i j}}\right)=\\
&=\frac{p_{i j}}{q_{i j} Z} \cdot\left(1+d_{k l}\right)^{-2}-\sum_{k \neq l} p_{k l} \frac{\left(1+d_{i j}\right)^{-2}}{Z}=\\
&=p_{i j}\left(1+d_{k l}\right)^{-1} - q_{i j}\left(1+d_{i j}\right)^{-1}=\\
&=\left(p_{i j} - q_{i j}\right)\left(1+d_{i j}\right)^{-1}.
\end{aligned}
\end{equation}

Inserting the above in \eqref{eq:dCdyi}, we obtain
\begin{equation}
\frac{\partial C}{\partial y_i}=2\sum_{j}\left(p_{i j} - q_{i j}\right)\left(1+d_{i j}\right)^{-1} \cdot \frac{\partial\left[k\left(y_i, y_i\right)-2 k\left(y_i, y_j\right)\right]}{\partial y_i},
\label{eq:e2e_ker_tsne_grad}
\end{equation}
where we used again the kernel trick in the reduced space
\begin{equation}
d_{i j}=\left\|\varphi\left(y_i\right)-\varphi\left(y_j\right)\right\|^2=k\left(y_i, y_i\right)-2 k\left(y_i, y_j\right)+k\left(y_j, y_j\right).
\end{equation}
Notice that all the optimizations are made through the lifted space.
By using different kernel functions (or the same function with different parameter values), we can endow the data space with different metrics, thus getting different forms of dimensional reduction.

Compared to the original t-SNE method, we introduce two main additional stages: we compute a kernel matrix and update the objective function with respect to the kernel distances. Notice that for the standard kernel t-SNE method, we only need to compute the kernel matrix once, for the data space; while for the end-to-end method, we also need to compute the kernel matrix of the reduced space at each iteration, since the reduced data points are updated.

As kernel function, we chose the Radial Basis Function (RBF) kernel $k(x, y) = \exp(-\gamma \|x - y\|^{2})$. We made this selection since this kernel maps the data into an infinite-dimensional feature space, being able to capture representative patterns relationships. Another reason why we chose this kernel for our experiments is its simplicity, since we only have a single hyperparameter to tune, the lengthscale $\gamma$.
Moreover, the kernel function is radially symmetric, depending only on the Euclidean distance between data points, being invariant and robust to any morphological transformations. 

\subsection{Possible improvements} 

The calculation of the kernel matrix is computationally costly and scales up with the dimensions of the used data set. This computation can be optimized by using different kernel approximations. A natural way to obtain this is the low-rank Nystrom approximation \cite{williams2000using}, \cite{yang2012nystrom}
\begin{equation}
    \tilde{K} = K_{nm} \cdot K_{mm}^{-1} \cdot K_{nm}^{\top}
\end{equation}
or the Random Fourier Features \cite{rahimi2007random} approximation for the kernel function $k(x, y) = \varphi_{r}(x)^{\top} \varphi_{r}(y)$ where
\begin{equation}
    \varphi_{r}(x) = [\cos(\omega_1^{\top}x + b_1), \cos(\omega_2^{\top}x + b_2), \cdots, \cos(\omega_r^{\top}x + b_r)]^{\top}.
\end{equation}
Other available improvements involve the creation of two spatial-hierarchy of the embeddings \cite{van2021efficient}, which are simultaneously traversed to approximate multiple N-body correlations, in a GPGPU-based environment.


\subsection{Preserving Local Structure}

To address the issue of overcrowding \cite{van2009learning}, a possible solution is to calculate pairwise similarities $q_{ij}$ using a heavy-tailed distribution in the latent space. By utilizing a heavy-tailed distribution, points far apart can be modeled as being even farther apart, which helps to eliminate the attractive forces responsible for overcrowding. A Student-t distribution, which is theoretically linked to the Gaussian distribution, is employed as the heavy-tailed distribution to measure these pairwise similarities. 

The low-dimensional mapping can be modified by replacing the Gaussian distribution
with the Student-t one, thus transforming \eqref{eq:q_kernel} into
\begin{equation}
q_{i j}=\frac{\left(1+\left\|\varphi\left(x_i\right)-\varphi\left(x_j\right)\right\|^2 / \alpha\right)^{-\frac{\alpha+1}{2}}}{\sum_{k \neq l}\left(1+\left\|\varphi\left(x_k\right)-\varphi\left(x_l\right)\right\|^2 / \alpha\right)^{-\frac{\alpha+1}{2}}},
\label{pls_low_probability}
\end{equation}
where $\alpha = \max(m - 1, 1)$ represents the degrees of freedom of the Student t-distribution.

\begin{figure*}[!ht]
\centerline{\includegraphics[width=\linewidth]{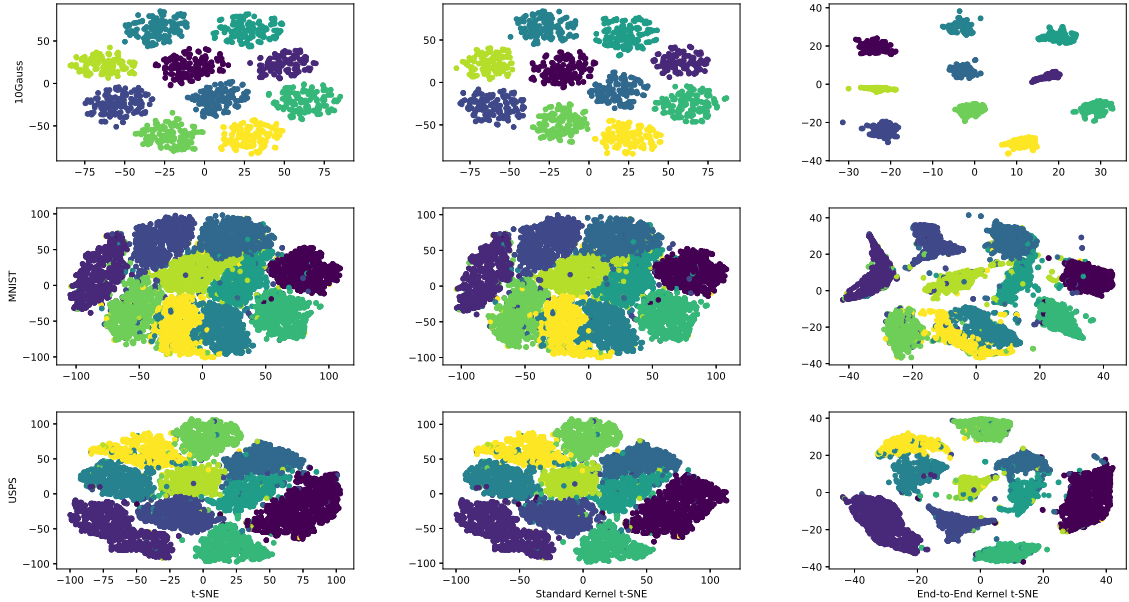}}
\caption{Dimensional reduction results}
\label{fig:dr_results}
\end{figure*}

For this adaptation, the update of the low-dimensional vectors is made with
\begin{equation}
\frac{\partial C}{\partial y_i}= \frac{\alpha + 1}{\alpha} \sum_{j}\left(p_{i j} - q_{i j}\right)\left(1+d_{i j} / \alpha\right)^{-1} \cdot \frac{\partial\left[k\left(y_i, y_i\right)-2 k\left(y_i, y_j\right)\right]}{\partial y_i}
\label{eq:alpha_e2e_ker_tsne}
\end{equation}
instead of \eqref{eq:e2e_ker_tsne_grad}.
We summarize the main steps of the proposed method in Algorithm \ref{alg:e2e_ker_tsne}.
We present here the end-to-end version of the Kernel t-SNE method, following the strategy of local structure-preserving.

\begin{algorithm}
\begin{algorithmic}[1]
\Require{$X$ - high dimensional data points, $m$ - target dimension, $p$ - perplexity, $\tau$ - learning rate, $\eta$ - momentum value, $n_\text{iter}$ - number of iterations, $k(u,v)$ - proposed kernel function}
\Ensure{$Y$ - low-dimensional representation of data}
\Procedure{Kernel tSNE}{$X$, $m$, $p$, $n_\text{iter}$, $k(x, y)$, $params$}
    \State Compute pairwise kernel distances $\hat{K} = \varphi(X)^{\top} \varphi(X)$ in the high-dimensional space from $X$
    \State Compute pairwise affinities $p_{j|i}$ based on pairwise kernel distances $\hat{K}_{ij}$ with \eqref{eq:pji_phi}
    \State Initialize low-dimensional points $Y^{(0)}$ with PCA
    \State Initialize momentum to $0$: $M^{(0)} \gets 0$
    \For{$t = 1$ to $n_\text{iter}$}
        \State Compute pairwise kernel distances $K = \varphi(Y)^{\top} \varphi(Y)$ in low-dimensional space from $Y^{(t-1)}$
        \State Compute pairwise affinities $q_{ij}$ based on pairwise kernel distances $K_{ij}$ with \eqref{pls_low_probability}
        \State Compute gradient: $G^{(t)}$ using \eqref{eq:alpha_e2e_ker_tsne}
        \State Update momentum: $M^{(t)} = \tau M^{(t-1)} - \eta G^{(t)}$
        \State Update low-dimensional points: $Y^{(t)} = Y^{(t-1)} + M^{(t)}$
    \EndFor
\EndProcedure
\caption{Kernel t-SNE Algorithm}
\label{alg:e2e_ker_tsne}
\end{algorithmic}
\end{algorithm}

\section{Implementation and results}
\label{sec:res}

In this section, we present the details of our implementation and the main results obtained with the proposed method Kernel t-SNE. The algorithm has been developed in Python, in the spirit of the scikit-learn framework \cite{scikit-learn}. Moreover, we had access to several implementations already available in the library. All our work is temporarily available in an anonymous \href{https://anonymous.4open.science/r/kernel-t-sne-4025/README.md}{Github repository}. All the experiments were made on a dedicated Desktop PC, with Ubuntu 20.04 as the operating system. We benefit from an i9 processor with 36 cores with a base frequency of 2.90 GHz (Max Turbo Frequency of 4.80 GHz), 256 GB RAM, and an NVIDIA RTX 3090 video card.

We used five different datasets for our experiments: COIL-20, Olivetti, Cardio, MNIST, and USPS. Additionally, we used a synthetic dataset proposed in \cite{rauber2016visualizing}; it consists of $2000$ data points of size $100$ samples over ten isotropic Gaussian distributions. We present here only the plots obtained for the two digits datasets. The COIL-20 dataset represents a set of images corresponding to 20 objects, under 72 different angles and orientations. The dataset sums up to a total number of 1440 images of size $32 \times 32$. The Olivetti dataset contains images of 40 different individuals of size $64 \times 64$ pixels. Each subject has 10 images in correspondence. The Cardiography (Cardio) dataset is available on the UCI machine learning repository. The samples represent features of cardiograms. The dataset counts a total number of 1831 samples of size 21. The MNIST dataset is a collection of handwritten digits available as grayscale images of size $28 \times 28$. It consists of $70.000$ registrations divided into $60.000$ for training and $10.000$ for testing. During our experiments, we used only batches of size $10.000$. The USPS dataset is an additional collection of handwritten digits, similar to MNIST, but with a lower resolution of $16 \times 16$. The total number of samples is $9298$. For both, we have digit labels from $0$ to $9$.

The Kernel t-SNE algorithm has been developed following the scikit-learn standard. The running stages of the algorithm are similar to the t-SNE. We introduced two additional steps: one for the computation of the kernel matrices alongside the corresponding gradients and the computation of the new objective function. For the computation of the kernel gradients in \eqref{eq:alpha_e2e_ker_tsne}, we used a finite difference approximation of the derivative of the scalar function. Taking into account that we follow a projection to the kernel space, we also added Kernel PCA as an initialization option for the low-dimensional space. However, in this paper, we present only the results obtained with the classical PCA initialization. 

\begin{table}[t]
\centering
\begin{tabular}{|l||*{4}{c|}}\hline
\backslashbox{Dataset}{Method} & \makebox[2.5em]{t-SNE} 
& \makebox[5em]{kernel t-SNE} & \makebox[6.5em]{e2e kernel t-SNE} \\\hline\hline
Cardio & 90.78\% & 90.82\% & \bf{92.81\%} \\\hline
COIL-20 & \bf{94.93\%} & 85.16\% & 87.28\% \\\hline
MNIST & 88.84\% & 88.89\% & \bf{90.04\%} \\\hline
Olivetti & 91.13\% & 90.45\% & \bf{91.92\%} \\\hline
USPS & 92.44\% & 92.45\% & \bf{93.31\%} \\\hline
\end{tabular}
\caption{Trustworthiness obtained on different datasets}
\label{tab:trustworthiness}
\end{table}

\begin{figure}[!ht]
\centerline{\includegraphics[width=7.5cm]{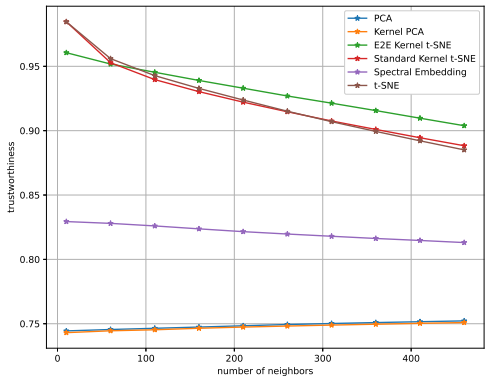}}
\caption{Trustworthiness evolution - MNIST dataset}
\label{fig:trustworthiness}
\end{figure}

To better understand the primary advantages of Kernel t-SNE, we computed projections into the 2D space and directly analyzed the results plot. We compare the results with other dimensional reduction methods such as PCA, Kernel PCA, Spectral Embedding, and the main competitor t-SNE. Another performance measurement is the trustworthiness \cite{venna2001neighborhood}, which indicates to what extent the local structure is retained. For each dataset, we followed a grid search procedure. We used only the RBF kernel with values $\gamma \in \{10^{-3}, 10^{-2}, 10^{-1}, 1, 10^{1}, 10^{2}, 10^{3}\}$. For the manifold learning, we used perplexity values $p \in \{10, 20, 30, 40, 50\}$. The early exaggeration was used with the default value of $12$. At the same time, the learning rate $\tau = \max(N / early\_exaggeration / 4,50)$ was set to auto, where $N$ is the number of samples, meaning to improve the trustworthiness value slightly, optimizations were performed over $n\_iter = 1000$ iterations, with the first $250$ used for early exaggeration. For the gradient descent stage, we used a momentum weight $\eta = 0.5$. For each dataset, we directly compared both versions of Kernel t-SNE (standard and end-to-end) with the t-SNE algorithm. In Figure \ref{fig:dr_results}, we can see that the kernelized version defines better boundaries between the clusters of different classes. Comparing the execution time of the three main methods we conclude that Kernel t-SNE has a running time similar to that of t-SNE. On the other hand, the End-to-End Kernel t-SNE is about three times slower due to the kernel matrix computation which is made on each iteration.

Moreover, the trustworthiness values demonstrate the good behavior of our algorithm, which can better retain the relationship between the local and global structure. In Table \ref{tab:trustworthiness}, we show the trustworthiness values obtained for each dataset considering a total number of $500$ neighbors. In general, the Standard Kernel t-SNE obtains similar 2D plots with t-SNE, managing to improve the trustworthiness value slightly. The end-to-end version produces the most notable results, getting better results in terms of visualization and trustworthiness. An exception to the proposed method is the COIL-20 dataset, for which the t-SNE method seems more suited. In Figure \ref{fig:trustworthiness}, we plot the evolution of the trustworthiness over different numbers of neighbors. The results shown there are the mean of three different subsampling tests. PCA and Kernel PCA have the worst results, although they tend to improve with the number of neighbors. Spectral Embedding obtains better results compared to the previous two. The best results are obtained by t-SNE and Kernel t-SNE.
The t-SNE method outperforms the other techniques for a small number of neighbors. On the other hand, the kernelized version scales up better with the number of neighbors beating all the other methods. During our experiments, we concluded that Kernel t-SNE does not need large perplexities since the global structure will be retained in the kernel space.

\section{Conclusions and future work}

This paper presents a new perspective on the t-SNE dimensional reduction algorithm through the kernel metric space, which can be used in high- and low-dimensional spaces. The proposed method can define better boundaries between clusters of different classes. The experiments demonstrate the good behavior of our method in terms of data visualization and analytics such as the trustworthiness measure, for which we obtain better results. The Kernel t-SNE can get better trustworthiness for a large number of neighbors. Compared to the original t-SNE, our method scales up better with the number of neighbors capable of retaining the local structure of data points.

We intend to improve our current work by introducing kernel approximation methods, such as Nystrom sampling or Random Fourier Features, to reduce the execution time. We also plan to propose a benchmark to illustrate the behavior of all parameters of the algorithm.

\newpage
\clearpage

\bibliographystyle{abbrv-doi}
\bibliography{main}
\end{document}